\pdfoutput=1
\documentclass[11pt, a4paper, nonumbering]{groundbench}
\usepackage[authoryear, sort&compress, round]{natbib}
\usepackage{ulem}
\usepackage{caption}
\usepackage{xspace}
\usepackage{multirow}
\usepackage{tcolorbox}
\usepackage{xltabular}
\usepackage{longtable}
\usepackage{hyperref}
\usepackage[capitalize,noabbrev]{cleveref} 
\usepackage{amsfonts}
\usepackage{amsmath}
\usepackage{amssymb}

\usepackage{mathtools}
\usepackage{physics}

\usepackage[bottom]{footmisc}
\microtypesetup{patch={item,verbatim,eqnum,toc}}

\usepackage{setspace}

\makeatletter
\def\@BTrule[#1]{%
  \ifx\longtable\undefined
    \let\@BTswitch\@BTnormal
  \else\ifx\hline\LT@hline
    \nobreak
    \let\@BTswitch\@BLTrule
  \else
     \let\@BTswitch\@BTnormal
  \fi\fi
  \global\@thisrulewidth=#1\relax
  \ifnum\@thisruleclass=\tw@\vskip\@aboverulesep\else
  \ifnum\@lastruleclass=\z@\vskip\@aboverulesep\else
  \ifnum\@lastruleclass=\@ne\vskip\doublerulesep\fi\fi\fi
  \@BTswitch}
\makeatother

\addto\extrasenglish{
}

 {\begin{list}{}%
         {\setlength{\leftmargin}{#1}}%
         \item[]%
 }
 {\end{list}}

\reportnumber{}
\correspondingauthor={Equal contribution.\\
Correspondence: \href{mailto:qizhangyang2000@gmail.com}{qizhangyang2000@gmail.com}, \href{mailto:bianzhonghan@gmail.com}{bianzhonghan@gmail.com}, \href{mailto:zhenran.w.1103@gmail.com}{zhenran.w.1103@gmail.com}}

\definecolor{gold}{HTML}{D4AF37}
\usepackage{algorithm}
\usepackage{algorithmic}
\definecolor{gbDoubao}{HTML}{3478BD}
\definecolor{gbKimi}{HTML}{F28E2B}
\definecolor{gbQwen}{HTML}{4E9D57}
\definecolor{gbGemini}{HTML}{E64B4E}
\definecolor{gbDiag}{HTML}{E87BA4}
\definecolor{gbBase}{HTML}{4A3AA7}
\definecolor{gbTruth}{HTML}{D03B3B}
\definecolor{gbHead}{HTML}{184F95}
\definecolor{gbTint}{HTML}{EDF3FD}
\definecolor{gbTintB}{HTML}{F5F4F1}
\definecolor{gbRule}{HTML}{9FB4CE}
\definecolor{gbInk2}{HTML}{52514E}

\hypersetup{colorlinks=true, citecolor=gbHead, linkcolor=gbHead, urlcolor=gbHead}
\titleformat{\section}{\large\bfseries\headingfont\color{gbHead}}{\color{gbHead}\thesection.}{0.5em}{#1}[]
\titleformat{name=\section,numberless}{\large\bfseries\headingfont\color{gbHead}}{}{0em}{#1}[]
\titleformat{\subsection}{\bfseries\color{gbHead}}{\color{gbHead}\thesubsection.}{0.5em}{#1}[]
\titleformat{\subsubsection}{\bfseries\itshape\color{gbHead}}{\color{gbHead}\thesubsubsection.}{0.5em}{#1}[]
\titleformat{\paragraph}[runin]{\bfseries\color{gbHead}}{}{0em}{#1}
\newcommand{\bench}{\textsc{GroundBench}\xspace}
\newcommand{\acc}[1]{\ensuremath{\mathrm{Acc}@#1}}
\newcommand{\thead}{\rowcolor{gbTint}}
\newcommand{\mdl}[2]{#2}
\colorlet{heat}{gbQwen}
\newcommand{\dgain}[1]{}
\newcommand{\dloss}[1]{}
\newcommand{\dwg}[1]{}
\newcommand{\dwgb}[1]{}
\newcommand{\dwgu}[1]{}
\newcommand{\ioushade}[2]{#2}
\newcommand{\accshade}[2]{#2}
\newcommand{\fmtprefix}[1]{}
\newcommand{\fmtsfx}[1]{}
\newcommand{\scorepair}[2]{#1/#2}
\newcommand{\bestscorepair}[2]{#1/\textbf{#2}}

\newcommand{\compactsubsection}[1]{\subsection{#1}}

\title{\centering \bench: Multi-Resolution Polygon Grounding\\ Exposes the Geometry Gap in Vision--Language Models}

\newsavebox{\gbfront}
\AtBeginDocument{\savebox{\gbfront}{%
  \setlength{\fboxsep}{6pt}%
  \colorbox{gbTint}{%
  \begin{tabular}{@{}r@{\ \ }l@{}}
  \small\textcolor{gbHead}{\textbf{Project page}} &
    \small\href{https://co-minder.github.io/Groundbench/}{\texttt{co-minder.github.io/Groundbench}}
  \end{tabular}}%
}}

\author[*]{%
  Zhonghan Bian\textsuperscript{*}, \; Zhenran Wang\textsuperscript{*}, \; Jinsong Li, \; Zhangyang Qi
  \par\vspace{11pt}\centerline{\usebox{\gbfront}}%
}

\renewcommand{\phi}{\varphi}

\renewcommand{\leq}{\leqslant}
\renewcommand{\geq}{\geqslant}

\renewcommand{\epsilon}{\varepsilon}
\renewcommand{\imath}{\mathrm{i}}

\newlength{\restsubwidth}
\newlength{\restsubheight}
\newlength{\restsubmoreheight}
\newcommand{\rest}[2]{%
        \settowidth{\restsubwidth}{\ensuremath{#2}}
        \settoheight{\restsubheight}{\ensuremath{{}_{#2}}}
        \ensuremath{{#1\hskip 0.5pt}_{\vrule\kern2pt\parbox[b][%
        4pt][b]{\the\restsubwidth}{%
                        \ensuremath{{}_{#2}}}}}
        }

\begin{abstract}
Bounding-box scores on RefCOCO-family grounding leave little room to distinguish frontier vision--language systems, yet boxes discard object shape. We introduce \bench, a matched benchmark that re-targets the same 1{,}500 image--expression--referent triples to exact-$N$ polygons at five vertex budgets. A fixed-denominator harness separately audits filled-region intersection over union (IoU) and legal-polygon completion. The strongest tested configuration reaches 88.2 box IoU and 97.1 accuracy at IoU $\ge .5$ (Acc@.5), versus 57.7 and 69.2 for direct polygons; because these headline scores use different references, we also compare direct polygons with predicted boxes rasterised against the same contour target, obtaining 57.7 versus 57.3 when pooled. Performance is non-monotone in $N$ and collapses at the densest budget, where legality failures compound residual geometric error. Qwen's thinking-setting contrast is the largest tested input-preserving configuration difference; under frozen templates, false spatial cues are more damaging than false colour cues, and target preference can remain high while contour tracing is poor. Alternate masks and a continuous-area scorer preserve the principal ordering. \bench therefore measures an operational output-geometry gap spanning localisation, boundary construction, serialisation, and topology, rather than latent boundary perception alone.
\end{abstract}

\begin{document}

\maketitle
% Pipeline (Figure 1): single-column, top of page 1.  ROLE: the one-glance overview.
\newcommand{\pipelinefloat}{%
\begin{figure}[t]
\centering
\includegraphics[width=0.909\linewidth]{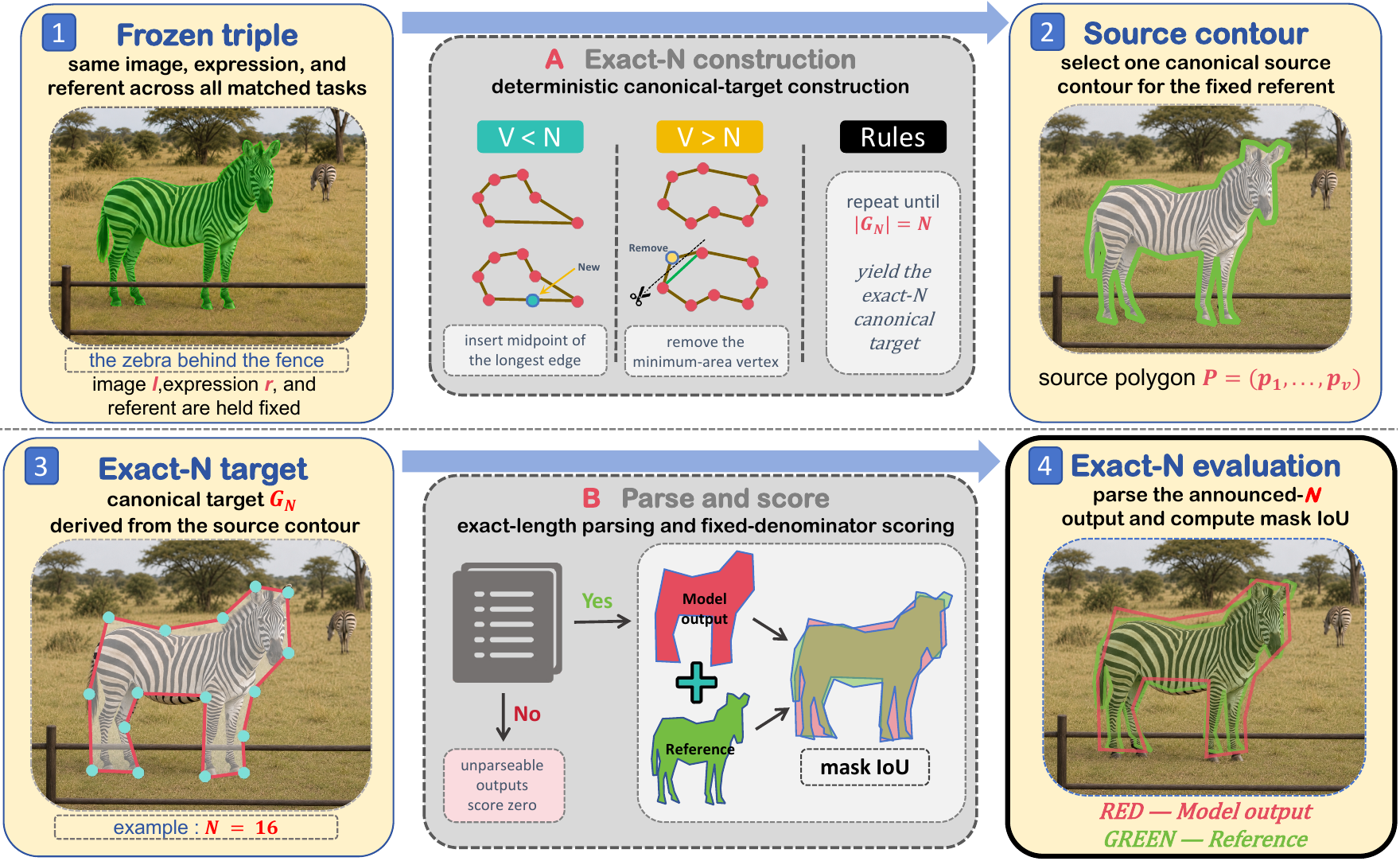}
\caption{\bench freezes each image--expression--referent triple, derives five exact-budget targets, and scores exact-length outputs with a deterministic rasteriser.}
\label{fig:method7}
\end{figure}
}

\section{Introduction}
\label{sec:intro7}

\pipelinefloat

Referring-expression grounding maps a phrase to an image region. RefCOCO-family benchmarks~\citep{RefCOCO,RefCOCOg}, built on COCO~\citep{COCO}, are evaluated mainly with 2D boxes. Frontier hosted large vision--language models (LVLMs) approach $90$ accuracy at IoU $\ge .5$ (\acc{.5})~\citep{qwen2.5-vl,InternVL} on our distractor-rich cohort, although we do not claim saturation of the full source distribution.

The limitation is the output contract. A four-number rectangle carries no shape, follows no boundary, and credits background interior, revealing little about point allocation along an object's contour. Dense masks restore shape but usually require a specialist decoder or external segmentation module~\citep{SAM}, coupling the multimodal decision to a separately trained geometric head.

% Main table first; Figure 2 follows as its visual summary.
\newcommand{\maintablefloat}{%
\begin{table}[t]
\centering
\setlength{\tabcolsep}{4pt}
\small
\resizebox{\textwidth}{!}{%
\begin{tabular}{l l c cc ccccc}
\toprule
\thead
$N$ & Configuration & $\mathbf{IoU}^{\mathbf{2}}$ & IoU$_{\mathrm{p}}$ & IoU$_{\mathrm{n}}$ & \acc{.1} & \acc{.3} & \acc{.5} & \acc{.7} & \acc{.9} \\
\cmidrule(l){2-10}
\raisebox{1.15ex}[0pt][0pt]{\small $N=8$} & \mdl{gbDoubao}{Doubao Seed-2.0-Pro} & \ioushade{10}{\fmtprefix{0}\scorepair{41.2}{41.2}} & \ioushade{11}{\scorepair{41.9}{41.9}} & \ioushade{11}{\scorepair{47.3}{47.3}} & \accshade{5}{\scorepair{75.0}{75.0}} & \accshade{5}{\scorepair{59.3}{59.3}} & \accshade{15}{\scorepair{43.3}{43.3}} & \accshade{21}{\scorepair{24.2}{24.2}} & \accshade{14}{\scorepair{1.7}{1.7}} \\
 & \mdl{gbKimi}{Kimi K2.7 Code} & \ioushade{5}{\fmtprefix{0}\scorepair{38.2}{38.2}} & \ioushade{5}{\scorepair{38.6}{38.6}} & \ioushade{5}{\scorepair{43.6}{43.6}} & \accshade{27}{\scorepair{\underline{83.8}}{83.8}} & \accshade{11}{\scorepair{63.1}{63.1}} & \accshade{5}{\scorepair{33.8}{33.8}} & \accshade{5}{\scorepair{10.1}{10.1}} & \accshade{5}{\scorepair{0.1}{0.1}} \\
 & \mdl{gbQwen}{Qwen 3.7 Plus} & \ioushade{23}{\fmtprefix{.4}\scorepair{\underline{48.2}}{48.4}} & \ioushade{24}{\scorepair{\underline{49.1}}{49.2}} & \ioushade{24}{\scorepair{\underline{55.4}}{55.6}} & \accshade{21}{\scorepair{81.3}{81.7}} & \accshade{24}{\scorepair{\underline{71.1}}{71.4}} & \accshade{29}{\scorepair{\underline{56.0}}{56.2}} & \accshade{28}{\scorepair{\underline{30.3}}{30.4}} & \accshade{15}{\scorepair{\underline{1.9}}{1.9}} \\
 & \mdl{gbGemini}{Gemini 3.5 Flash} & \ioushade{50}{\fmtprefix{0}\bestscorepair{\textbf{62.6}}{62.6}} & \ioushade{50}{\bestscorepair{\textbf{63.6}}{63.6}} & \ioushade{50}{\bestscorepair{\textbf{71.7}}{71.7}} & \accshade{50}{\bestscorepair{\textbf{93.0}}{93.0}} & \accshade{50}{\bestscorepair{\textbf{87.8}}{87.8}} & \accshade{50}{\bestscorepair{\textbf{76.0}}{76.0}} & \accshade{50}{\bestscorepair{\textbf{48.8}}{48.8}} & \accshade{50}{\bestscorepair{\textbf{8.0}}{8.0}} \\
\cmidrule(l){2-10}
\raisebox{1.15ex}[0pt][0pt]{\small $N=16$} & \mdl{gbDoubao}{Doubao Seed-2.0-Pro} & \ioushade{21}{\fmtprefix{0}\scorepair{46.8}{46.8}} & \ioushade{21}{\scorepair{46.8}{46.8}} & \ioushade{21}{\scorepair{48.7}{48.7}} & \accshade{5}{\scorepair{78.7}{78.7}} & \accshade{14}{\scorepair{65.1}{65.1}} & \accshade{24}{\scorepair{51.5}{51.5}} & \accshade{26}{\scorepair{\underline{33.4}}{33.4}} & \accshade{19}{\scorepair{\underline{3.5}}{3.5}} \\
 & \mdl{gbKimi}{Kimi K2.7 Code} & \ioushade{5}{\fmtprefix{.1}\scorepair{37.3}{37.3}} & \ioushade{5}{\scorepair{37.3}{37.3}} & \ioushade{5}{\scorepair{38.8}{38.8}} & \accshade{18}{\scorepair{\underline{82.3}}{82.3}} & \accshade{5}{\scorepair{60.0}{60.0}} & \accshade{5}{\scorepair{32.9}{32.9}} & \accshade{5}{\scorepair{10.1}{10.1}} & \accshade{5}{\scorepair{0.1}{0.1}} \\
 & \mdl{gbQwen}{Qwen 3.7 Plus} & \ioushade{25}{\fmtprefix{.3}\scorepair{\underline{49.7}}{49.9}} & \ioushade{25}{\scorepair{\underline{49.7}}{49.8}} & \ioushade{25}{\scorepair{\underline{51.6}}{51.8}} & \accshade{15}{\scorepair{81.4}{81.7}} & \accshade{26}{\scorepair{\underline{71.5}}{71.8}} & \accshade{31}{\scorepair{\underline{58.1}}{58.3}} & \accshade{26}{\scorepair{33.3}{33.4}} & \accshade{14}{\scorepair{2.2}{2.2}} \\
 & \mdl{gbGemini}{Gemini 3.5 Flash} & \ioushade{50}{\fmtprefix{.2}\bestscorepair{\textbf{64.7}}{64.9}} & \ioushade{50}{\bestscorepair{\textbf{64.7}}{64.9}} & \ioushade{50}{\bestscorepair{\textbf{67.2}}{67.3}} & \accshade{50}{\bestscorepair{\textbf{90.6}}{90.8}} & \accshade{50}{\bestscorepair{\textbf{84.6}}{84.8}} & \accshade{50}{\bestscorepair{\textbf{76.0}}{76.2}} & \accshade{50}{\bestscorepair{\textbf{59.8}}{59.9}} & \accshade{50}{\bestscorepair{\textbf{11.0}}{11.0}} \\
\cmidrule(l){2-10}
\raisebox{1.15ex}[0pt][0pt]{\small $N=24$} & \mdl{gbDoubao}{Doubao Seed-2.0-Pro} & \ioushade{16}{\fmtprefix{0}\scorepair{43.4}{43.4}} & \ioushade{16}{\scorepair{43.3}{43.3}} & \ioushade{16}{\scorepair{44.1}{44.1}} & \accshade{5}{\scorepair{76.9}{76.9}} & \accshade{8}{\scorepair{60.0}{60.0}} & \accshade{19}{\scorepair{45.6}{45.6}} & \accshade{24}{\scorepair{29.1}{29.1}} & \accshade{16}{\scorepair{\underline{2.8}}{2.8}} \\
 & \mdl{gbKimi}{Kimi K2.7 Code} & \ioushade{5}{\fmtprefix{0}\scorepair{35.9}{35.9}} & \ioushade{5}{\scorepair{35.8}{35.8}} & \ioushade{5}{\scorepair{36.5}{36.5}} & \accshade{20}{\scorepair{81.8}{81.8}} & \accshade{5}{\scorepair{57.9}{57.9}} & \accshade{5}{\scorepair{29.7}{29.7}} & \accshade{5}{\scorepair{7.2}{7.2}} & \accshade{5}{\scorepair{0.3}{0.3}} \\
 & \mdl{gbQwen}{Qwen 3.7 Plus} & \ioushade{25}{\fmtprefix{.4}\scorepair{\underline{49.1}}{49.3}} & \ioushade{25}{\scorepair{\underline{49.1}}{49.3}} & \ioushade{25}{\scorepair{\underline{49.9}}{50.1}} & \accshade{21}{\scorepair{\underline{82.1}}{82.5}} & \accshade{26}{\scorepair{\underline{71.3}}{71.6}} & \accshade{29}{\scorepair{\underline{56.6}}{56.8}} & \accshade{27}{\scorepair{\underline{32.4}}{32.5}} & \accshade{11}{\scorepair{1.7}{1.7}} \\
 & \mdl{gbGemini}{Gemini 3.5 Flash} & \ioushade{50}{\fmtprefix{.6}\bestscorepair{\textbf{65.9}}{66.3}} & \ioushade{50}{\bestscorepair{\textbf{65.8}}{66.2}} & \ioushade{50}{\bestscorepair{\textbf{67.0}}{67.4}} & \accshade{50}{\bestscorepair{\textbf{92.0}}{92.6}} & \accshade{50}{\bestscorepair{\textbf{87.2}}{87.7}} & \accshade{50}{\bestscorepair{\textbf{79.6}}{80.1}} & \accshade{50}{\bestscorepair{\textbf{58.2}}{58.6}} & \accshade{50}{\bestscorepair{\textbf{11.0}}{11.1}} \\
\cmidrule(l){2-10}
\raisebox{1.15ex}[0pt][0pt]{\small $N=32$} & \mdl{gbDoubao}{Doubao Seed-2.0-Pro} & \ioushade{25}{\fmtprefix{.1}\scorepair{44.4}{44.4}} & \ioushade{25}{\scorepair{44.3}{44.3}} & \ioushade{25}{\scorepair{44.8}{44.8}} & \accshade{5}{\scorepair{77.1}{77.2}} & \accshade{18}{\scorepair{61.2}{61.2}} & \accshade{26}{\scorepair{46.8}{46.8}} & \accshade{30}{\scorepair{30.6}{30.6}} & \accshade{18}{\scorepair{\underline{3.0}}{3.0}} \\
 & \mdl{gbKimi}{Kimi K2.7 Code} & \ioushade{5}{\fmtprefix{.7}\scorepair{33.7}{34.0}} & \ioushade{5}{\scorepair{33.7}{33.9}} & \ioushade{5}{\scorepair{34.1}{34.3}} & \accshade{20}{\scorepair{78.5}{79.1}} & \accshade{5}{\scorepair{55.3}{55.7}} & \accshade{5}{\scorepair{26.8}{27.0}} & \accshade{5}{\scorepair{6.5}{6.6}} & \accshade{5}{\scorepair{0.2}{0.2}} \\
 & \mdl{gbQwen}{Qwen 3.7 Plus} & \ioushade{33}{\fmtprefix{.3}\scorepair{\underline{48.6}}{48.8}} & \ioushade{33}{\scorepair{\underline{48.6}}{48.8}} & \ioushade{33}{\scorepair{\underline{49.1}}{49.2}} & \accshade{50}{\scorepair{\textbf{81.4}}{81.7}} & \accshade{39}{\scorepair{\underline{71.3}}{71.5}} & \accshade{37}{\scorepair{\underline{56.7}}{56.9}} & \accshade{30}{\scorepair{\underline{30.9}}{31.0}} & \accshade{13}{\scorepair{1.9}{1.9}} \\
 & \mdl{gbGemini}{Gemini 3.5 Flash} & \ioushade{50}{\fmtprefix{8.6}\bestscorepair{\textbf{57.8}}{63.3}} & \ioushade{50}{\bestscorepair{\textbf{57.8}}{63.2}} & \ioushade{50}{\bestscorepair{\textbf{58.4}}{63.9}} & \accshade{44}{\bestscorepair{\underline{80.8}}{88.4}} & \accshade{50}{\bestscorepair{\textbf{76.4}}{83.6}} & \accshade{50}{\bestscorepair{\textbf{69.4}}{75.9}} & \accshade{50}{\bestscorepair{\textbf{49.8}}{54.5}} & \accshade{50}{\bestscorepair{\textbf{10.0}}{10.9}} \\
\cmidrule(l){2-10}
\raisebox{1.15ex}[0pt][0pt]{\small $N=64$} & \mdl{gbDoubao}{Doubao Seed-2.0-Pro} & \ioushade{47}{\fmtprefix{9.7}\scorepair{37.6}{41.6}} & \ioushade{47}{\scorepair{37.5}{41.6}} & \ioushade{47}{\scorepair{37.8}{41.8}} & \accshade{50}{\scorepair{\textbf{69.3}}{76.7}} & \accshade{46}{\scorepair{\underline{54.6}}{60.5}} & \accshade{41}{\scorepair{39.5}{43.8}} & \accshade{31}{\scorepair{21.3}{23.6}} & \accshade{15}{\scorepair{\underline{1.7}}{1.9}} \\
 & \mdl{gbKimi}{Kimi K2.7 Code} & \ioushade{5}{\fmtprefix{19.2}\scorepair{24.4}{30.2}} & \ioushade{5}{\scorepair{24.4}{30.2}} & \ioushade{5}{\scorepair{24.5}{30.4}} & \accshade{20}{\scorepair{58.3}{72.1}} & \accshade{5}{\scorepair{38.5}{47.6}} & \accshade{5}{\scorepair{18.9}{23.4}} & \accshade{5}{\scorepair{5.1}{6.3}} & \accshade{5}{\scorepair{0.0}{0.0}} \\
 & \mdl{gbQwen}{Qwen 3.7 Plus} & \ioushade{50}{\fmtprefix{14.3}\scorepair{\textbf{38.5}}{44.9}} & \ioushade{50}{\scorepair{\textbf{38.5}}{44.9}} & \ioushade{50}{\scorepair{\textbf{38.7}}{45.2}} & \accshade{49}{\scorepair{\underline{68.8}}{80.3}} & \accshade{50}{\scorepair{\textbf{56.3}}{65.7}} & \accshade{46}{\scorepair{\underline{42.5}}{49.6}} & \accshade{33}{\scorepair{\underline{22.7}}{26.5}} & \accshade{11}{\scorepair{1.1}{1.2}} \\
 & \mdl{gbGemini}{Gemini 3.5 Flash} & \ioushade{47}{\fmtprefix{40.6}\bestscorepair{\underline{37.7}}{63.4}} & \ioushade{47}{\bestscorepair{\underline{37.7}}{63.4}} & \ioushade{47}{\bestscorepair{\underline{37.9}}{63.7}} & \accshade{5}{\bestscorepair{52.8}{88.9}} & \accshade{33}{\bestscorepair{49.4}{83.2}} & \accshade{50}{\bestscorepair{\textbf{44.8}}{75.4}} & \accshade{50}{\bestscorepair{\textbf{33.6}}{56.6}} & \accshade{50}{\bestscorepair{\textbf{7.6}}{12.8}} \\
\cmidrule(l){2-10}
\raisebox{1.15ex}[0pt][0pt]{\small Pooled} & \mdl{gbDoubao}{Doubao Seed-2.0-Pro} & \ioushade{22}{\fmtprefix{2.0}\scorepair{42.7}{43.5}} & \ioushade{22}{\scorepair{42.8}{43.6}} & \ioushade{21}{\scorepair{44.5}{45.4}} & \accshade{5}{\scorepair{75.4}{76.9}} & \accshade{15}{\scorepair{60.1}{61.3}} & \accshade{24}{\scorepair{45.3}{46.3}} & \accshade{26}{\scorepair{27.7}{28.3}} & \accshade{17}{\scorepair{\underline{2.5}}{2.6}} \\
 & \mdl{gbKimi}{Kimi K2.7 Code} & \ioushade{5}{\fmtprefix{4.0}\scorepair{33.9}{35.3}} & \ioushade{5}{\scorepair{34.0}{35.4}} & \ioushade{5}{\scorepair{35.5}{37.0}} & \accshade{16}{\scorepair{76.9}{80.1}} & \accshade{5}{\scorepair{55.0}{57.2}} & \accshade{5}{\scorepair{28.4}{29.6}} & \accshade{5}{\scorepair{7.8}{8.1}} & \accshade{5}{\scorepair{0.1}{0.2}} \\
 & \mdl{gbQwen}{Qwen 3.7 Plus} & \ioushade{29}{\fmtprefix{3.2}\scorepair{\underline{46.8}}{48.4}} & \ioushade{29}{\scorepair{\underline{47.0}}{48.5}} & \ioushade{29}{\scorepair{\underline{48.9}}{50.5}} & \accshade{30}{\scorepair{\underline{79.0}}{81.6}} & \accshade{32}{\scorepair{\underline{68.3}}{70.5}} & \accshade{33}{\scorepair{\underline{54.0}}{55.7}} & \accshade{29}{\scorepair{\underline{29.9}}{30.9}} & \accshade{13}{\scorepair{1.8}{1.8}} \\
 & \mdl{gbGemini}{Gemini 3.5 Flash} & \ioushade{50}{\fmtprefix{10.0}\bestscorepair{\textbf{57.7}}{64.2}} & \ioushade{50}{\bestscorepair{\textbf{57.9}}{64.3}} & \ioushade{50}{\bestscorepair{\textbf{60.4}}{67.1}} & \accshade{50}{\bestscorepair{\textbf{81.8}}{90.9}} & \accshade{50}{\bestscorepair{\textbf{77.1}}{85.6}} & \accshade{50}{\bestscorepair{\textbf{69.2}}{76.8}} & \accshade{50}{\bestscorepair{\textbf{50.0}}{55.6}} & \accshade{50}{\bestscorepair{\textbf{9.5}}{10.6}} \\
\midrule
 & \textit{Mean} & \textbf{45.3/47.9} & \textbf{45.4/48.0} & \textbf{47.3/50.0} & \textbf{78.3/82.4} & \textbf{65.1/68.7} & \textbf{49.2/52.1} & \textbf{28.9/30.7} & \textbf{3.5/3.8} \\
\bottomrule
\end{tabular}%
}
\caption{\bench filled-region results ($0$--$100$). Every metric cell shows fixed\,/\,parseable-only scoring under IoU$^{2}$. IoU$_{\mathrm p}$ uses the source part and IoU$_{\mathrm n}$ is fidelity-normalised. Pooled and Mean average budgets and configurations; bold/underline mark best/second.}
\label{tab:main7}
\end{table}
}
\maintablefloat

We therefore study an intermediate representation: expressive enough to expose shape, but compact and textual enough to be generated through the same interface used for general vision--language interaction. A fixed-budget polygon couples three abilities. The model must resolve the expression among same-category distractors, infer the visible boundary, and serialise a closed ordered sequence with exactly the requested number of vertices. Increasing $N$ offers more representational capacity but also lengthens the response and creates more opportunities for duplicated points, crossings, shortcuts, and accumulated drift. Repeating the question across budgets exposes where added bandwidth stops helping.

We introduce \bench, a multi-resolution polygon grounding benchmark for hosted LVLMs. Given an image, expression, and vertex count $N$, a model returns exactly $N$ normalised points. Five deterministic targets, $N\in\{8,16,24,32,64\}$, are derived from one selected annotation contour and fixed before evaluation (\cref{fig:method7}), so image, phrase, and referent stay constant while only the answer surface changes; all five files share the same 1{,}500 UIDs for exact-intersection paired analysis. The cohort draws 500 questions each from RefCOCO, RefCOCO+, and RefCOCOg, with frozen metadata for distractors ($D$), contour complexity ($Q$), relative scale ($RS$), semantic group, phrase length, and a composite difficulty label. We report two fixed-denominator axes---filled-region utility and legal-polygon completion---each scoring an unparseable response zero.

Evaluating four frontier configurations across all five budgets reveals four findings. \textbf{(F1)}~Even the strongest configuration leaves a large gap between box localisation and polygon production. \textbf{(F2)}~Accuracy is non-monotone in $N$ and collapses at the densest budget despite a more faithful reference contour. \textbf{(F3) The Qwen thinking-setting contrast is the largest input-preserving difference:} max-thinking consistently beats thinking-off, while removing the image is comparably damaging and resolution or greyscale changes matter less. \textbf{(F4)}~Injected cues act asymmetrically---wrong spatial hints hurt more than wrong colours, while truthful hints show no reliable gain.

\paragraph{Contributions.}
First, we formulate fixed-vertex polygon grounding as a direct text-generation task and release 1{,}500 questions with five matched targets per referent. Second, we expose benchmark composition rather than one opaque average: source, semantic group, $D$, $Q$, $RS$, phrase length, and a reproducible difficulty label are frozen per question. Third, we compare strong hosted configurations under one frozen protocol, pairing every cross-budget claim on exact question intersections. Fourth, we separate formatting, target preference, and contour tracing, probe them with a frozen 13-condition grid (reasoning, visual evidence, resolution, colour, cue truth, decoding, repeatability), and verify the conclusions under alternate masks and an independent backend.

\section{Related Work}
\label{sec:related7}

% Figure 2 follows the main table and visualises its budget-wise trends.
\begin{figure}[t]
\centering
\includegraphics[width=0.92\linewidth]{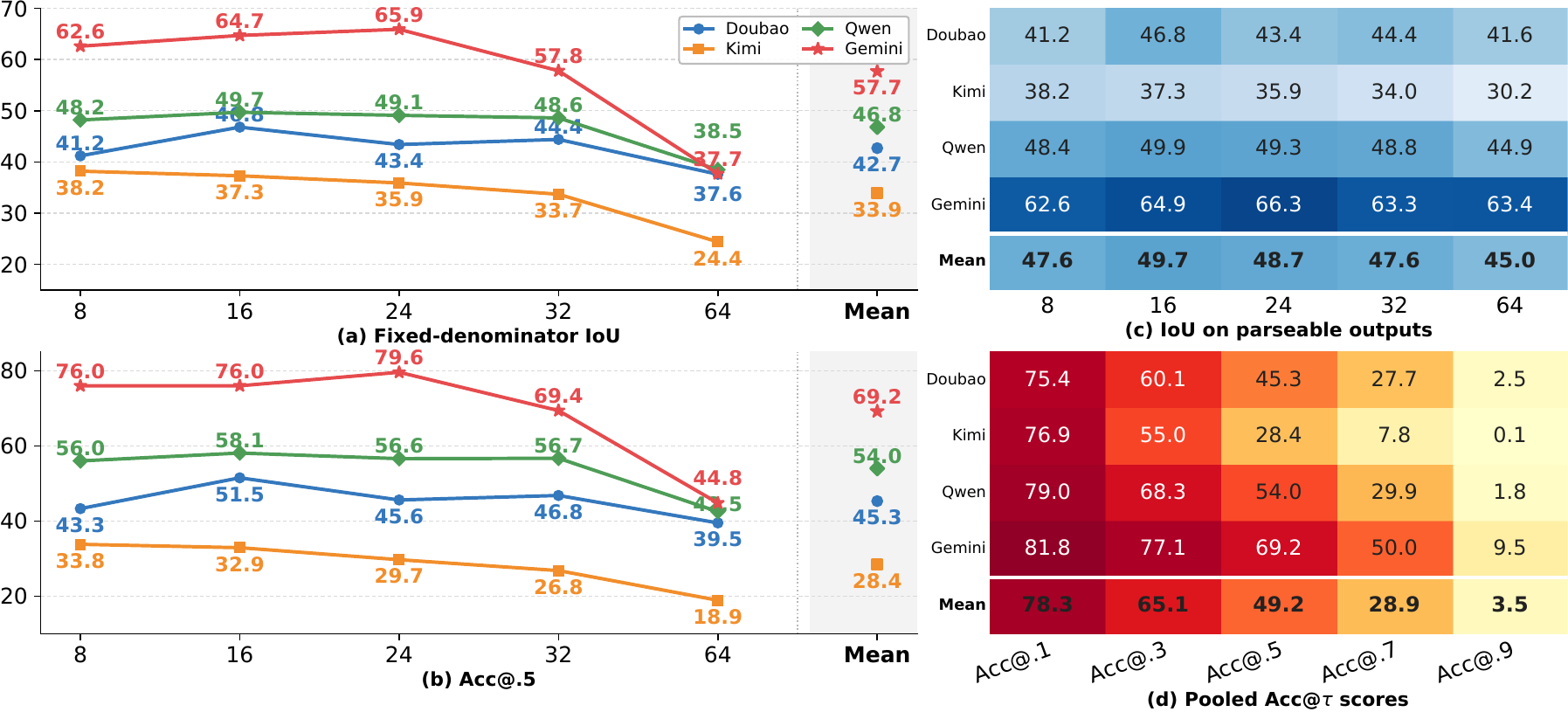}
\caption{Summary of \cref{tab:main7}: fixed IoU$^{2}$, Acc@.5, parseable-only IoU, and pooled threshold accuracy.}
\label{fig:curve7}
\end{figure}

\paragraph{Referring and polygon grounding.}
RefCOCO-family benchmarks pair language with COCO instances~\citep{ReferItGame,RefCOCO,RefCOCOg,COCO}, with later variants targeting segmentation masks, open phrases, or multiple objects~\citep{MDETR,PhraseCut,GRES}. Sequence models cast localisation as coordinate generation~\citep{Seqtr,PVD}, PolyFormer~\citep{PolyFormer} generates variable-length polygons for referring segmentation, and interactive contour models predict boundaries as vertex sequences~\citep{PolygonRNN,PolygonRNNpp,CurveGCN,DeepSnake}---all trained specialist decoders. \bench is complementary: it audits general-purpose hosted LVLMs with no task-specific training, an announced output budget, and the same referents across a multi-resolution curve.

\paragraph{Grounded multimodal models.}
Generalist systems increasingly emit boxes, points, or region references in text~\citep{kosmos2,Shikra,Ferret,VisionLLM,Qwen2-VL,qwen2.5-vl}, while grounded assistants connect conversation to dense region machinery~\citep{LISA,GroundingGPT,NExT-Chat,GLaMM}. Recent work emits segmentation directly and extends grounding beyond RGB~\citep{Qwen3VLSeg,GenSegR1,RGBTGround}; \bench instead audits unmodified hosted APIs. Neither a short box nor a learned mask head establishes that a chat decoder can maintain a long cyclic boundary. Fixed-$N$ polygons make allocation, exact-length compliance, and topology directly auditable between box grounding and specialist segmentation.

\noindent\textbf{Coordinates as an output language.} Once localisation is emitted as text~\citep{pix2seq}, coordinate vocabulary, sequence length, and ordering become part of the task. Fixing $N$ and reusing each referent across budgets prevents adaptive shortcuts and separates representational capacity from delivery. Spatial suites report persistent perceptual and relational weaknesses~\citep{BLINK,EyesWideShut,VSR,SpatialVLM}; \bench instead requires a dense, metrically scored geometric artefact rather than a categorical judgement.

\noindent\textbf{Polygon evaluation.} Rasterising an ordered boundary into a scored region involves explicit choices about disconnected parts, self-intersections, clipping, and discrete integration. We therefore freeze one simple official policy and rescore the same predictions under alternate reference masks and continuous-area intersection, bounding whether any conclusion depends on one evaluator.

\noindent\textbf{Saturation-driven benchmark design.} A recurring response to a saturated benchmark is to re-target the same task to a harder version rather than to a new domain: ImageNet$\rightarrow$ImageNet-V2 and ReaL \citep{ImageNet,ImageNetV2,ImageNetReaL}, GLUE$\rightarrow$SuperGLUE \citep{GLUE,SuperGLUE}, MMLU$\rightarrow$MMLU-Pro \citep{MMLU,MMLUPro}, MATH$\rightarrow$AIME \citep{MATH}. \bench follows this pattern for 2D referring-expression grounding---bounding box $\rightarrow$ multi-resolution polygon on the same RefCOCO source. The matched design isolates the harder output contract from changes in the image, expression, and referent distribution.

\section{The \bench Benchmark}
\label{sec:method7}

\subsection{Task and Canonical Targets}
For image $I$ of width $W$ and height $H$, expression $r$, and announced budget $N$, the model outputs $\widehat P=(p_1,\ldots,p_N)$ with $p_i=(x_i,y_i)$. The prompt requests one line of $2N$ integers in $\{0,\ldots,999\}$. Filled scoring maps finite coordinates to $(xW/999,\,yH/999)$ and rasterises their clipped cyclic fill; legality scoring requires distinct in-range integer vertices and a simple positive-area polygon, scoring failures zero. Start point and traversal direction are unrestricted because filled-mask IoU is invariant to cyclic shifts and reversal. A sensitivity audit reconstructs targets after shifting the sampling origin through eight offsets spaced at $12.5\%$ of the contour.

When an annotation contains multiple polygon parts, construction selects the part with the most listed vertices. This affects $148/1{,}500$ UIDs; within them, the selected part covers $71.76\%$ of the all-parts union on average and is also the largest-area part in $85.14\%$. \Cref{alg:target} derives its canonical target $G_N$, and \cref{fig:method7} illustrates the process: a sparse contour is upsampled by repeatedly inserting the midpoint of the currently longest edge, and a dense contour is reduced by Visvalingam--Whyatt simplification~\citep{VisvalingamWhyatt}, which iteratively removes the vertex with the smallest neighbour-triangle area until exactly $N$ vertices remain; ties take the lowest index, keeping construction deterministic.

This frozen representation is rescored against the unreduced part, all-parts union, alternate starting phases, and continuous-area integration. The parser accepts the last explicit or bracketed list only when it contains exactly $2N$ finite values. Output allowances are nonbinding: the longest legal answer is about $520$ characters, versus a tightest allowance of $1{,}280$ tokens.

% ablation block (cue + thinking tables, with radar) declared here, mid-3.2,
% so it is encountered while p4 typesets and claims the p5 top.
% BEGIN INLINED FROM table/ablation_block.tex
% Ablation block: two-table row in the permitted 9pt \small size.
\begin{table}[t]
  \begin{minipage}[t]{0.495\linewidth}\centering
    {\small\setlength{\tabcolsep}{2pt}\renewcommand{\arraystretch}{1.34}% BEGIN INLINED FROM table/abl_prompt_cues.tex
% Table 2 body: prompt cues, image evidence, and the B_cue reference are
% separated with booktabs rules.  Wrong-cue (W) corner deltas use bold for
% the largest magnitude per column and underline for the second.
% Upper-right gray digit (\fmtsfx) = parse shortfall % on every cell, 0 shown
% as 0 (source: final-ablation-grid-v2 cell_summary; B_cue parses 100%).
\begin{tabular}[b]{@{}l|ccc|c@{}}
\toprule[1.5pt]
\thead
Cue & $N=8$ & $N=16$ & $N=24$ & Mean \\
\midrule
S (true spatial) & $45.5\fmtsfx{0}\dloss{2.3}$ & $47.2\fmtsfx{0}\dloss{2.0}$ & $48.5\fmtsfx{0}\dgain{0.3}$ & $\mathbf{47.1}\fmtsfx{0}\dloss{1.4}$ \\
C (true colour) & $45.3\fmtsfx{.7}\dloss{2.6}$ & $46.5\fmtsfx{1.0}\dloss{2.8}$ & $45.4\fmtsfx{1.5}\dloss{2.8}$ & $\mathbf{45.7}\fmtsfx{1.1}\dloss{2.7}$ \\
SC (both true) & $44.7\fmtsfx{.7}\dloss{3.1}$ & $46.9\fmtsfx{.2}\dloss{2.4}$ & $45.8\fmtsfx{1.8}\dloss{2.4}$ & $\mathbf{45.8}\fmtsfx{.9}\dloss{2.6}$ \\
WC (wrong colour) & $38.5\fmtsfx{.5}\dwg{9.3}$ & $39.4\fmtsfx{0}\dwg{9.9}$ & $39.6\fmtsfx{2.5}\dwg{8.6}$ & $\mathbf{39.2}\fmtsfx{1.0}\dwg{9.3}$ \\
WS (wrong spatial) & $22.6\fmtsfx{.7}\dwgu{25.3}$ & $24.6\fmtsfx{1.2}\dwgu{24.7}$ & $25.8\fmtsfx{1.7}\dwgu{22.5}$ & $\mathbf{24.3}\fmtsfx{1.2}\dwgu{24.1}$ \\
WSC (both wrong) & $20.6\fmtsfx{.3}\dwgb{27.2}$ & $22.6\fmtsfx{.8}\dwgb{26.7}$ & $23.2\fmtsfx{3.2}\dwgb{25.0}$ & $\mathbf{22.1}\fmtsfx{1.4}\dwgb{26.3}$ \\
\midrule
No image & $14.1\fmtsfx{0}\dloss{34.6}$ & $13.2\fmtsfx{0}\dloss{35.9}$ & $13.3\fmtsfx{0}\dloss{35.5}$ & $\mathbf{13.5}\fmtsfx{0}\dloss{35.3}$ \\
Half resolution & $42.4\fmtsfx{0}\dloss{6.1}$ & $41.6\fmtsfx{0}\dloss{7.7}$ & $42.5\fmtsfx{0}\dloss{6.7}$ & $\mathbf{42.2}\fmtsfx{0}\dloss{6.8}$ \\
Greyscale & $46.8\fmtsfx{0}\dloss{1.3}$ & $48.2\fmtsfx{0}\dloss{1.6}$ & $47.5\fmtsfx{.4}\dloss{1.5}$ & $\mathbf{47.5}\fmtsfx{.1}\dloss{1.5}$ \\
\midrule
$B_{\mathrm{cue}}$ (none) & $\mathbf{47.8}\fmtsfx{0}$ & $\mathbf{49.3}\fmtsfx{0}$ & $\mathbf{48.2}\fmtsfx{0}$ & $\mathbf{48.4}\fmtsfx{0}$ \\
\bottomrule[1.5pt]
\end{tabular}

% END INLINED FROM table/abl_prompt_cues.tex
}
    \caption{\textbf{Prompt cues and image evidence.} Fixed IoU ($\times100$); $B_{\mathrm{cue}}$ and image controls are same-UID matched.}
    \label{tab:abl-cue}
  \end{minipage}
  \hfill
  \begin{minipage}[t]{0.495\linewidth}\centering
    % \unskip + trailing % kill the trailing spaces after the tabular: with them
    % present the box+space exceed \hsize and the line breaker emits a phantom
    % second line, pushing this caption one baseline (~11pt) below its table.
    {\small\setlength{\tabcolsep}{2pt}\renewcommand{\arraystretch}{1.06}% BEGIN INLINED FROM table/abl_input_decoding.tex
% Table 3 body: thinking depth + decoding/repeatability vs B_full (max-think).
% Each thinking level (max-think shown redundantly at the top, then 4096, off)
% carries a second, gray row with its parseable-only IoU; the parse shortfall sits
% in the fixed row's upper-right superscript (\fmtsfx).  Base B_full is last.
% Booktabs rules distinguish thinking depth, decoding/repeatability, and base.
\begin{tabular}[b]{@{}l|ccc|c@{}}
\toprule[1.1pt]
\thead
Condition & $N=8$ & $N=16$ & $N=24$ & Mean \\
\midrule
\multicolumn{5}{@{}l@{}}{\textit{Thinking depth}}\\
Max thinking & $48.2\fmtsfx{.4}$ & $49.7\fmtsfx{.3}$ & $49.1\fmtsfx{.4}$ & $\mathbf{49.0}\fmtsfx{.4}$ \\
\hspace{0.7em}\textit{parse.} & \textcolor{gbInk2}{48.4} & \textcolor{gbInk2}{49.9} & \textcolor{gbInk2}{49.3} & \textcolor{gbInk2}{49.2} \\
4,096 tokens & $48.0\fmtsfx{1.4}\dloss{0.2}$ & $49.0\fmtsfx{.4}\dloss{0.7}$ & $47.4\fmtsfx{4.6}\dloss{1.7}$ & $\mathbf{48.1}\fmtsfx{2.1}\dloss{0.9}$ \\
\hspace{0.7em}\textit{parse.} & $\textcolor{gbInk2}{48.7}\dgain{0.3}$ & $\textcolor{gbInk2}{49.2}\dloss{0.7}$ & $\textcolor{gbInk2}{49.7}\dgain{0.4}$ & $\textcolor{gbInk2}{49.2}$ \\
Thinking off & $20.5\fmtsfx{24.1}\dloss{27.7}$ & $14.7\fmtsfx{21.2}\dloss{35.0}$ & $17.0\fmtsfx{12.7}\dloss{32.1}$ & $\mathbf{17.4}\fmtsfx{19.3}\dloss{31.6}$ \\
\hspace{0.7em}\textit{parse.} & $\textcolor{gbInk2}{27.0}\dloss{21.4}$ & $\textcolor{gbInk2}{18.7}\dloss{31.2}$ & $\textcolor{gbInk2}{19.4}\dloss{29.9}$ & $\textcolor{gbInk2}{21.7}\dloss{27.5}$ \\
\midrule
\multicolumn{5}{@{}l@{}}{\textit{Decoding \& repeatability}}\\
Temperature 0 & $48.2\fmtsfx{.4}$ & $49.7\fmtsfx{.3}$ & $49.1\fmtsfx{.4}$ & $\mathbf{49.0}\fmtsfx{.4}$ \\
Temperature 0.5 & $47.2\fmtsfx{1.8}\dloss{1.0}$ & $48.6\fmtsfx{.4}\dloss{1.1}$ & $50.9\fmtsfx{1.4}\dgain{1.7}$ & $\mathbf{48.9}\fmtsfx{1.2}\dloss{0.1}$ \\
Temperature 1.0 & $45.9\fmtsfx{.8}\dloss{2.3}$ & $50.5\fmtsfx{.6}\dgain{0.8}$ & $49.2\fmtsfx{2.4}\dgain{0.1}$ & $\mathbf{48.5}\fmtsfx{1.3}\dloss{0.5}$ \\
Independent repeat & $46.7\fmtsfx{.8}\dloss{1.5}$ & $49.1\fmtsfx{.4}\dloss{0.6}$ & $49.9\fmtsfx{.6}\dgain{0.7}$ & $\mathbf{48.6}\fmtsfx{.6}\dloss{0.4}$ \\
\midrule
$B_{\mathrm{full}}$ (Qwen max-think) & $\mathbf{48.2}\fmtsfx{.4}$ & $\mathbf{49.7}\fmtsfx{.3}$ & $\mathbf{49.1}\fmtsfx{.4}$ & $\mathbf{49.0}\fmtsfx{.4}$ \\
\bottomrule[1.1pt]
\end{tabular}

% END INLINED FROM table/abl_input_decoding.tex
\unskip}%
    \caption{\textbf{Thinking, decoding, and repeatability.} Fixed IoU ($\times100$); italics report parseable-only scores.}
    \label{tab:abl-params}
  \end{minipage}
\end{table}

\begin{algorithm}[t]
\caption{Canonical target construction $G_N$}
\label{alg:target}
\begin{algorithmic}[1]
\REQUIRE source contour $C=(c_1,\ldots,c_V)$, budget $N$
\WHILE{$|C|<N$}
  \STATE $k \gets \arg\max_j \lVert c_{j+1}-c_j \rVert$ \COMMENT{longest edge, cyclic}
  \STATE insert $(c_k+c_{k+1})/2$ after $c_k$
\ENDWHILE
\WHILE{$|C|>N$}
  \STATE $k \gets \arg\min_j \tfrac{1}{2}\lvert(c_{j-1}-c_j)\times(c_{j+1}-c_j)\rvert$
  \STATE remove $c_k$ \COMMENT{smallest neighbour-triangle area}
\ENDWHILE
\STATE \textbf{return} $G_N \gets C$
\end{algorithmic}
\end{algorithm}

\subsection{Cohort Construction and Rationale}
Three frozen stages select 1{,}500 unique image--referent pairs: 500 each from RefCOCO, RefCOCO+, and RefCOCOg, one per image. Six frozen groups cover the 80 COCO categories. Distractor bins $D0/D1/D2/D3{+}$ denote zero, one, two, or at least three other non-crowd target-category instances; the release contains $705/276/519$ questions in $D1/D2/D3{+}$ and none in $D0$. Contour complexity is $Q=\ell^2/(4\pi A)$, where $\ell$ and $A$ are source-contour perimeter and area; relative scale $RS$ is target-box area divided by image area. $Q1/Q2/Q3$ and $RS1/RS2/RS3$ are low/middle/high bins under globally frozen thresholds. Cohort sampling preceded all model outputs.

For phrase word count $L$ and source vertex count $V$, a deterministic composite
\begin{align}
s={}&w_D+w_Q+w_{RS}+\mathbf{1}[L\!\geq\!6]+\mathbf{1}[L\!\geq\!10]\nonumber\\
&+\mathbf{1}[V\!\geq\!25]+\mathbf{1}[V\!\geq\!50],
\end{align}
with $w_D=(0,0,1,2)$ for $(D0,D1,D2,D3{+})$, $w_Q=(0,1,2)$ for $(Q1,Q2,Q3)$, and $w_{RS}=(2,1,0)$ for $(RS1,RS2,RS3)$. This yields $436$ easy ($s\!\leq\!2$), $554$ medium ($3\!\leq\!s\!\leq\!4$), and $510$ hard ($s\!\geq\!5$) questions. The components are correlated, so we report both the individual axes and the summary label.

% END INLINED FROM table/ablation_block.tex

% Single-column visual summary of the controlled ablations.
\begingroup
\setlength{\parskip}{0pt}
\setlength{\intextsep}{4pt}

\compactsubsection{Scoring}
All scores in this paper are reported on a $100$-point scale. Let $R(\cdot)$ denote rasterisation onto the original canvas and $\mathbf{1}[\rho_i]$ mark a successful exact-length parse; on failure, the product below is defined as zero without evaluating $R(\widehat P_i)$. For threshold $\tau$, the per-question score and official aggregates over planned cohort $S$ with parsed subset $S_{\mathrm{p}}$ are
\begin{equation}
z_i = \mathbf{1}[\rho_i]\,\operatorname{IoU}\!\big(R(\widehat P_i),R(G_{N,i})\big),
\label{eq:zi}
\end{equation}
\begin{equation}
\mathrm{IoU}_{\mathrm{fix}} = \tfrac{100}{|S|}\!\sum_{i\in S} z_i,\qquad
\mathrm{IoU}_{\mathrm{parse}} = \tfrac{100}{|S_{\mathrm{p}}|}\!\sum_{i\in S_{\mathrm{p}}}\! z_i,
\label{eq:agg}
\end{equation}
\begin{equation}
\acc{\tau} = \tfrac{100}{|S|}\!\sum_{i\in S}\mathbf{1}[z_i\ge\tau].
\label{eq:acc}
\end{equation}
The fixed denominator makes every planned question count, and \cref{eq:agg} pairs every fixed value with its parseable-only counterpart because conditioning on successful output can make a low-coverage configuration appear geometrically strong. Filled scoring asks whether an emitted sequence still covers a useful region; the separate legality-enforced audit retains the fixed denominator but assigns zero unless the parsed coordinates form a legal polygon. The alternate-mask columns of \cref{tab:main7} replace the canonical target in \cref{eq:zi}: with $A_i$ the unreduced selected source part, set $z_i^{\mathrm p}=0$ on a parse failure and otherwise
{\small
\begin{equation}
z^{\mathrm p}_i=\mathrm{IoU}(R(\widehat P_i),A_i),
\quad \mathrm{IoU}_{\mathrm n}=100\,\mathrm{IoU}_{\mathrm p}/C_N .
\label{eq:variants}
\end{equation}
}
\begin{sloppypar}
where IoU$_{\mathrm{p}}$ aggregates $z^{\mathrm{p}}_i$ as in \cref{eq:agg} and the construction fidelity $C_N={}$\allowbreak$100\operatorname{mean}_i \operatorname{IoU}(R(G_{N,i}),A_i)$ is $88.6/96.2/98.3/99.0/99.4$ for $N=8/16/24/32/64$. Higher $C_N$ means less approximation loss. IoU$_{\mathrm{n}}$ is the descriptive representation-fidelity ratio $100\,\mathrm{IoU}_{\mathrm p}/C_N$; it is neither a probability nor a mathematical upper bound and may exceed $100$. Superscripts retain the recovery history: IoU$^{0}$ is the primary pass, IoU$^{1}$ follows one parse-failure-triggered stateless retry, and IoU$^{2}$ follows up to two; all headline scores, contrasts, and remaining main-table metrics use the uniform IoU$^{2}$ view; best-effort is reserved for explicitly labelled diagnostics defined in the Setup.
\end{sloppypar}
\endgroup

% Full-width selection-versus-tracing table, declared with its diagnostic.
\begin{table}[t]
\centering
{\small\setlength{\tabcolsep}{1.6pt}%
\resizebox{\textwidth}{!}{%
\begin{tabular}{@{}l *{15}{>{\centering\arraybackslash}p{9.4mm}}@{}}
\toprule
\thead
 & \multicolumn{3}{c}{$N{=}8$} & \multicolumn{3}{c}{$N{=}16$} & \multicolumn{3}{c}{$N{=}24$} & \multicolumn{3}{c}{$N{=}32$} & \multicolumn{3}{c}{$N{=}64$} \\
\cmidrule(lr){2-4}\cmidrule(lr){5-7}\cmidrule(lr){8-10}\cmidrule(lr){11-13}\cmidrule(lr){14-16}
\thead
Config & Sel & Con & IoU & Sel & Con & IoU & Sel & Con & IoU & Sel & Con & IoU & Sel & Con & IoU \\
\midrule
\mdl{gbDoubao}{Doubao} & $76.9$ & $53.9$ & $\mathbf{41.2}$ & $80.5$ & $57.7$ & $\mathbf{46.8}$ & $78.7$ & $54.5$ & $\mathbf{43.4}$ & $80.0$ & $54.9$ & $\mathbf{44.4}$ & $71.2$ & $52.0$ & $\mathbf{37.6}$ \\
\mdl{gbKimi}{Kimi} & $85.5$ & $44.6$ & $\mathbf{38.2}$ & $84.7$ & $43.3$ & $\mathbf{37.3}$ & $83.5$ & $42.1$ & $\mathbf{35.9}$ & $80.9$ & $40.8$ & $\mathbf{33.7}$ & $60.7$ & $38.5$ & $\mathbf{24.4}$ \\
\mdl{gbQwen!55!gbInk2}{Qwen-off} & $60.9$ & $33.8$ & $\mathbf{20.5}$ & $62.6$ & $23.4$ & $\mathbf{14.7}$ & $68.7$ & $24.2$ & $\mathbf{17.0}$ & $71.3$ & $25.6$ & $\mathbf{18.8}$ & $3.8$ & $25.8$ & $\mathbf{1.0}$ \\
\mdl{gbQwen}{Qwen-max} & $85.2$ & $57.5$ & $\mathbf{48.2}$ & $86.5$ & $57.6$ & $\mathbf{49.7}$ & $85.7$ & $57.4$ & $\mathbf{49.1}$ & $85.9$ & $56.7$ & $\mathbf{48.6}$ & $70.5$ & $54.1$ & $\mathbf{38.5}$ \\
\mdl{gbGemini}{Gemini} & $94.2$ & $68.1$ & $\mathbf{62.6}$ & $92.4$ & $70.8$ & $\mathbf{64.7}$ & $93.4$ & $71.3$ & $\mathbf{65.9}$ & $86.0$ & $70.1$ & $\mathbf{59.5}$ & $62.2$ & $67.7$ & $\mathbf{41.9}$ \\
\midrule
\textit{Mean} & $\mathbf{80.5}$ & $\mathbf{51.6}$ & $\mathbf{42.1}$ & $\mathbf{81.3}$ & $\mathbf{50.6}$ & $\mathbf{42.7}$ & $\mathbf{82.0}$ & $\mathbf{49.9}$ & $\mathbf{42.2}$ & $\mathbf{80.8}$ & $\mathbf{49.6}$ & $\mathbf{41.0}$ & $\mathbf{53.7}$ & $\mathbf{47.6}$ & $\mathbf{28.7}$ \\
\bottomrule
\end{tabular}%
}}
\caption{Selection/tracing by budget in the best-effort view. Sel is target preference, Con is target IoU conditional on Sel, and IoU uses the fixed denominator.}
\label{tab:abl-select}
\end{table}

\subsection{Selection and Tracing Diagnostics}
Diagnostics compare each parsed prediction with the intended all-parts target mask $U_i$ and every same-category distractor mask $D_{i,d}$, where $d$ indexes those distractors; unparsed outputs receive $\mathrm{Sel}_i=0$. Unique target preference and conditional tracing for parsed outputs are
{\small
\begin{align}
\mathrm{Sel}_i&=\mathbf{1}\!\left[
\mathrm{IoU}(R(\widehat P_i),U_i)>\max\nolimits_d\mathrm{IoU}(R(\widehat P_i),D_{i,d})
\right],\label{eq:sel}\\
\mathrm{Con}
  &=100\operatorname{mean}\{\operatorname{IoU}(R(\widehat P_i),U_i):\mathrm{Sel}_i=1\}.
\label{eq:con}
\end{align}
}
Thus Sel records a preference event only when target overlap exceeds every distractor overlap, and Con measures tracing conditional on that event. This is not verified instance detection: it has no absolute overlap threshold, confidence ranking, margin, or cross-category matching. We additionally audit duplicate points, zero area, self-intersection, alternate reference regions, and continuous-area IoU on already-legal polygons. Same-UID comparisons preserve exact question pairing, and the image interventions use exact same-UID controls.

\section{Experiments}
\label{sec:experiments7}

\subsection{Setup}

Four hosted configurations---Doubao, Kimi, Qwen-max, and Gemini-high---answer 1{,}500 questions at five budgets, with Qwen-off as a same-endpoint ablation. Generations were collected 10--17 July 2026; images kept aspect ratio with the longest side capped at 1{,}024 pixels. This is a purposive hosted-system snapshot, not a probability sample or architecture-level comparison, so every result is attributed to the complete labelled configuration. Headline IoU$^{2}$ locks the first parseable output over one primary call plus two stateless retries (IoU$^{0}$ is the one-call view), while the separately labelled best-effort diagnostic adds one authorised Gemini recovery and never enters headline rankings.

% Table 5 is declared before the main-results prose so that Table 4 and Table 5
% remain on consecutive pages without forced float placement.
% BEGIN INLINED FROM table/difficulty_heatmatrix.tex
\begin{table}[t]
\centering
{\small\setlength{\tabcolsep}{2.4pt}%
\resizebox{\textwidth}{!}{%
\begin{tabular}{@{}l|ccc|ccc|ccc|ccc|cccccc@{}}
\toprule
\thead
& \multicolumn{3}{c|}{\textbf{Difficulty}} & \multicolumn{3}{c|}{\textbf{Distr.} $D$} & \multicolumn{3}{c|}{\textbf{Compl.} $Q$} & \multicolumn{3}{c|}{\textbf{Scale} $RS$} & \multicolumn{6}{c}{\textbf{Semantic group}} \\
\cmidrule(lr){2-4}\cmidrule(lr){5-7}\cmidrule(lr){8-10}\cmidrule(lr){11-13}\cmidrule(lr){14-19}
\thead
Cfg. & Ea & Me & Ha & $D1$ & $D2$ & $D3{+}$ & $Q1$ & $Q2$ & $Q3$ & $RS1$ & $RS2$ & $RS3$ & An. & Food & Ind. & Out. & Per. & Veh. \\
\midrule
\mdl{gbDoubao}{Doubao} & $50.7$ & $43.0$ & $35.4$ & $45.9$ & $40.5$ & $39.4$ & $46.5$ & $44.5$ & $35.9$ & $37.0$ & $42.9$ & $49.8$ & $42.8$ & $41.9$ & $39.9$ & $37.6$ & $\underline{46.7}$ & $44.9$ \\
\mdl{gbKimi}{Kimi} & $42.5$ & $33.7$ & $26.8$ & $36.3$ & $33.1$ & $31.0$ & $39.6$ & $33.8$ & $26.5$ & $29.3$ & $34.2$ & $39.6$ & $32.3$ & $34.1$ & $35.0$ & $30.6$ & $33.9$ & $36.5$ \\
\mdl{gbQwen!55!gbInk2}{Qwen-off} & $17.5$ & $13.7$ & $12.5$ & $15.9$ & $13.8$ & $12.7$ & $15.3$ & $14.0$ & $13.6$ & $12.6$ & $14.6$ & $16.5$ & $17.3$ & $13.9$ & $11.9$ & $12.8$ & $15.6$ & $14.1$ \\
\mdl{gbQwen}{Qwen-max} & $\underline{55.6}$ & $\underline{46.4}$ & $\underline{39.8}$ & $\underline{50.3}$ & $\underline{45.7}$ & $\underline{42.6}$ & $\underline{52.2}$ & $\underline{46.9}$ & $\underline{39.6}$ & $\underline{42.9}$ & $\underline{47.3}$ & $\underline{51.5}$ & $\underline{46.7}$ & $\underline{48.1}$ & $\underline{45.5}$ & $\underline{42.7}$ & $46.4$ & $\underline{50.3}$ \\
\mdl{gbGemini}{Gemini} & $\mathbf{65.1}$ & $\mathbf{58.1}$ & $\mathbf{51.0}$ & $\mathbf{61.8}$ & $\mathbf{57.6}$ & $\mathbf{52.3}$ & $\mathbf{63.4}$ & $\mathbf{57.8}$ & $\mathbf{50.2}$ & $\mathbf{54.2}$ & $\mathbf{58.8}$ & $\mathbf{61.4}$ & $\mathbf{63.4}$ & $\mathbf{59.3}$ & $\mathbf{53.6}$ & $\mathbf{59.4}$ & $\mathbf{57.4}$ & $\mathbf{61.3}$ \\
\midrule
\textit{Mean} & $\mathbf{46.3}$ & $\mathbf{39.0}$ & $\mathbf{33.1}$ & $\mathbf{42.0}$ & $\mathbf{38.2}$ & $\mathbf{35.6}$ & $\mathbf{43.4}$ & $\mathbf{39.4}$ & $\mathbf{33.2}$ & $\mathbf{35.2}$ & $\mathbf{39.6}$ & $\mathbf{43.8}$ & $\mathbf{40.5}$ & $\mathbf{39.5}$ & $\mathbf{37.2}$ & $\mathbf{36.6}$ & $\mathbf{40.0}$ & $\mathbf{41.4}$ \\
\bottomrule
\end{tabular}%
}}
\caption{Pooled fixed-IoU slices. Ea/Me/Ha are easy/medium/hard; An./Ind./Out./Per./Veh. are animal/indoor/outdoor/person/vehicle. Difficulty, $D$, $Q$, and $RS$ use IoU$^{2}$; semantic groups use best-effort. Bold/underline mark best/second.}
\label{tab:difficulty7}
\label{tab:semantic}
\end{table}
% END INLINED FROM table/difficulty_heatmatrix.tex

\subsection{Main Results}
\Cref{tab:main7,fig:curve7} report fixed and parseable-only IoU, alternate-reference diagnostics, and threshold accuracy under uniform IoU$^{2}$. The two headline findings of the introduction emerge here together with a rank-stability pattern.

\textbf{(F1) A large, unresolved geometry gap.} Gemini leads, followed by Qwen-max, Doubao, and Kimi. Because headline box IoU uses the ground-truth box while contour IoU uses $G_N$, their separation is contextual rather than latent geometric loss. In the matched control, each predicted rectangle is uniformly sampled along its four edges into an exact-$N$ polygon scored against $G_N$; pooled Gemini direct and box-derived IoU are $57.7$ and $57.3$. Thresholds distinguish Kimi's low-budget object finding from Gemini's stronger conditional geometry, while both lose high-$N$ reliability.

\textbf{(F2) Non-monotone in $N$, collapse at $64$.} Results peak at moderate budgets and fall at the densest one despite increasing reference fidelity, and the decline persists when extra vertices only subdivide edges without changing the filled target---consistent with allocation, ordering, and topology burdens beyond parsing. At $N=64$, Gemini's filled IoU stays $37.7$ while legal-polygon coverage/IoU fall to $7.5/4.2$, so the paired axes separate residual region utility from legal completion. Eight starting phases preserve the configuration ordering and the high-budget reversal. Thus $N$ trades representation against delivery rather than defining one-dimensional difficulty.

\textbf{Rank stability.} $\acc{.5}$ preserves one ordering across the four main configurations and all five budgets, whereas fixed IoU has one Qwen--Gemini inversion at $N=64$. Lower thresholds can induce further cross-budget changes, so one threshold at one budget underdetermines model quality. We therefore report multiple thresholds across the full budget curve.

% Wide two-panel ablation figure, placed with the section that discusses it.
\begingroup
\setlength{\intextsep}{4pt}
\begin{figure}[tb]
  \centering
  \includegraphics[width=\columnwidth]{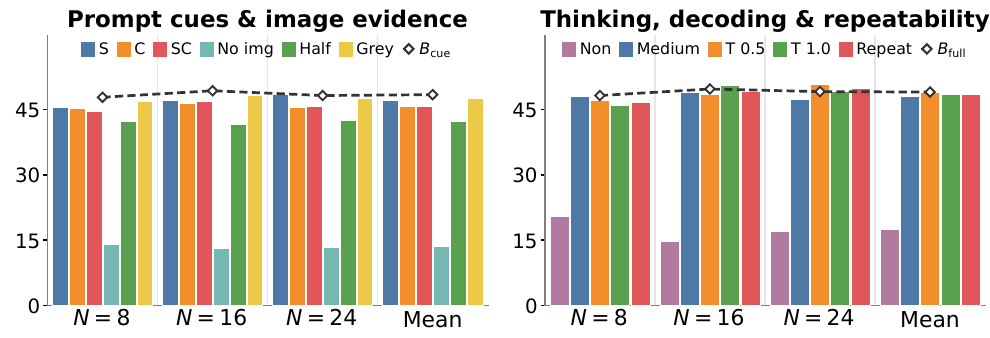}
  \caption{Controlled fixed-IoU ablations. Diamonds mark controls; circles mark matched-image controls.}
  \label{fig:ablradar}
\end{figure}
\endgroup

\subsection{Ablation Study}
\label{sec:ablation7}
\textbf{Controlled interventions.} \Cref{tab:abl-cue,tab:abl-params,fig:ablradar} report a frozen 13-condition Qwen grid (1{,}500 UIDs per budget, except matched 500-UID no-image, half-resolution, and greyscale cohorts, with Gemini replicating the image interventions). \textbf{(F3)}~Reasoning configuration is the largest input-preserving effect: max-thinking exceeds thinking-off at every budget, through both broader parse coverage and better geometry. Removing the image is comparably damaging, while halving resolution or removing colour cost far less (evidence~$\gg$~resolution~$\gg$~colour). \textbf{(F4)}~Cues act asymmetrically: wrong spatial hints hurt more than wrong colours and combine sub-additively, whereas truthful cues give no reliable gain---diagnosing susceptibility to conflicting language rather than an intrinsic spatial/colour ranking. Thinking-budget, temperature, and a repeat run all stay second-order.

\textbf{Diagnostics and robustness.} The selection/tracing split (\cref{tab:abl-select}) shows target preference staying high until the densest budget while conditional contour tracing is the persistent differentiator, so one IoU should not be read as either ability alone. Slicing the answers by composite difficulty, its components, and semantic group (\cref{tab:difficulty7}) gives consistent easy-to-hard declines, with matched parseable-only losses implicating conditional geometry rather than format coverage. Alternate reference masks and a continuous-area backend shift five-budget means by at most $1.1$ points without changing the ordering. As the cohort concentrates on same-class distractors within COCO/RefCOCO and the systems are time-stamped hosted configurations, \bench is a controlled output audit, not an architecture-wide claim.

\section{Conclusion}
\label{sec:conclusion7}
We introduce \bench, a matched benchmark that converts referring-expression grounding from boxes into fixed-budget polygon production. Aligned targets, deterministic scoring, frozen metadata, and parsing, topology, preference, and tracing diagnostics support controlled study of representation and delivery. Across matched budgets, richer contours eventually give way to parsing and topology failures, and the diagnostics localise where performance is lost---so the curve helps identify whether a new interface fixes parsing, contour construction, or topology as the boundary densifies. The paired region-utility and legality axes expose distinct failures, giving a common substrate for comparing future vision--language or tool-assisted systems. Coordinate emission is an audit stress test, not a prescribed interface; mask heads, tool calls, and compressed contours are valid alternatives.

\bibliographystyle{abbrvnat}
\bibliography{main}

\end{document}